\documentclass[letterpaper, 10 pt, conference]{ieeeconf}  

\IEEEoverridecommandlockouts                              

\usepackage{graphicx} 
\usepackage{amsmath} 
\usepackage{amssymb}  
\def\bh{\mathbf{h}}
\def\bS{\mathbf{S}}
\def\bL{\mathbf{L}}
\def\mN{\mathcal{N}}
\DeclareMathOperator*{\argmin}{arg\,min}
\newcommand{\bra}[1]{\left(#1\right)}
\newcommand{\brac}[1]{\left\{#1\right\}}
\newcommand{\norm}[1]{\Vert #1\Vert}
\title{\LARGE \bf
An End-to-End System for Crowdsourced 3d Maps for Autonomous Vehicles: The Mapping Component
}
\author{
Onkar Dabeer$^{1}$, Radhika Gowaikar$^{1}$, Slawomir K. Grzechnik$^{1}$, Mythreya J. Lakshman$^{1}$, Gerhard Reitmayr$^{2}$ \and Kiran Somasundaram$^{1}$, Ravi Teja Sukhavasi$^{1}$, Xinzhou Wu$^{1}$
\thanks{$^{1}$Qualcomm Technologies, Inc.}%
\thanks{$^{2}$Qualcomm Austria Research Center GmbH}%
\thanks{$^{\ast}$order in which authors' names appear does not signify contribution}
}
\begin{document}
\maketitle
\thispagestyle{empty}
\pagestyle{empty}
\begin{abstract}
Autonomous vehicles rely on precise high definition (HD) 3d maps for navigation. This paper presents the mapping component of an end-to-end system for crowdsourcing precise 3d maps with semantically meaningful landmarks such as traffic signs (6 dof pose, shape and size) and traffic lanes (3d splines). The system uses consumer grade parts, and in particular, relies on a single front facing camera and a consumer grade GPS. Using real-time sign and lane triangulation on-device in the vehicle, with offline sign/lane clustering across multiple journeys and offline Bundle Adjustment across multiple journeys in the backend, we construct maps with mean absolute accuracy at sign corners of less than 20 cm from 25 journeys. To the best of our knowledge, this is the first end-to-end HD mapping pipeline in global coordinates in the automotive context using cost effective sensors.
\end{abstract}
\section{INTRODUCTION}
In the past decade several autonomous vehicle prototypes have been demonstrated in highway and urban scenarios. The next decade is expected to be marked by commercial deployment of autonomous vehicles with different levels of autonomy (L1-L5, e.g., see \cite{SAEJ3016}). Many current prototypes rely on precise 3D maps \cite{Levinson2011, Bertha2014} and it is expected that most of the commercial designs will also leverage such maps. These maps serve two main purposes:
\begin{itemize}
\item The landmarks/features in the map can be associated with camera/lidar landmarks/features seen by the ego vehicle and this association can be used to precisely locate the ego vehicle in the map. An accuracy of 10 cm localization in the map is desired for navigation.
\item The map also provides prior information about the static environment, which can far exceed the range of the vehicle sensors. In other words, the map is also a sensor – for example, it can warn about a sharp turn 50 m ahead even though the camera view is occluded by a truck in front of the ego vehicle.
\end{itemize}
While there are a lot of classical techniques for 3d reconstruction of point clouds (e.g., Ch 7 \cite{Szeliski2010}), for the second reason above, maps with semantically meaningful objects, such as traffic signs and lanes, are of interest. Moreover, the map is a live object. Roads and associated signage changes frequently and reliable means of updating the map with new elements and discarding old elements is desired. For a globally scalable solution, crowdsourcing where vehicles using the map also contribute to its update is an attractive option. However, in contrast to specialized vehicle fleets with high grade expensive equipment used by many map makers, crowdsourcing inherently needs a map making solution based on consumer grade equipment to keep the system cost low. The key contribution of this paper is an end-to-end mapping solution based on a single front facing automotive grade 1MP RGB camera, consumer grade GPS, IMU (accelerometer and gyro), a Qualcomm Snapdragon  820A SoC in the vehicle, and a backend mapping server. In our current experiments with real data, we get mean absolute accuracy of less than 20cm and relative error of about 15 cm for traffic sign corners from 25 journeys. Our system comprises of several modules.
\begin{itemize}
\item Positioning: Precise camera 6dof pose estimation in the world coordinates runs in real-time on a Snapdragon 820A. We get lane level accuracy, which allows reliable association of landmarks across multiple journeys. In this paper, we do not address this module.
\item Perception: Precise detection and tracking of traffic sign corners and traffic lanes runs in real-time on a Qualcomm Snapdragon 820A. The perception module is described in \cite{Heeseok2017}.
\item Single-journey Triangulation: The positioning and perception outputs are processed on the same Snapdragon 820A SoC for real-time triangulation of traffic signs and lanes. We do a full 3d reconstruction: 6dof pose for signs and 3d splines for lanes. The triangulation outputs, related positioning and perception data are shipped over a commercial LTE link to the backend mapping server. 
\item Multi-journey Association/Clustering: In the backend mapping server, we cluster the triangulation outputs from multiple journeys across different days and vehicles. In particular, we identify the landmarks to be included in the map and their association with positioning and perception observations from individual journeys.
\item Multi-journey Bundle Adjustment: Finally, we generate the map by joint optimization of camera poses, sign poses and lane spline parameters to minimize a robust cost function. 
\end{itemize}
In this paper, we cover the last three modules above. Specifically, we focus on a) the rationale behind our design choices for triangulation, multi-journey processing, b) specific modules such as road normal estimation, which are critical for good performance, c) reporting performance on real data collected in San Diego, CA, USA. Our main message is that we can build precise 3d maps with consumer grade equipment and few tens of journeys.
The rest of the paper is organized as follows. Section \ref{sec:arch} describes the mapping component system architecture and the rationale behind the design choices. Section \ref{sec:sj} describes the triangulation pipeline and the offline multi-journey processing is described in Section \ref{sec:mj}. Section \ref{sec:results} summarizes our experiments on real world data. 
\section{SYSTEM ARCHITECTURE}
\label{sec:arch}
Our system architecture is primarily driven by modularity, which allows different teams to iterate rapidly. In this section, we describe the mapping pipeline and the motivation behind some of our design choices.

\begin{figure*}[thpb]
\centering
\includegraphics[scale=0.6]{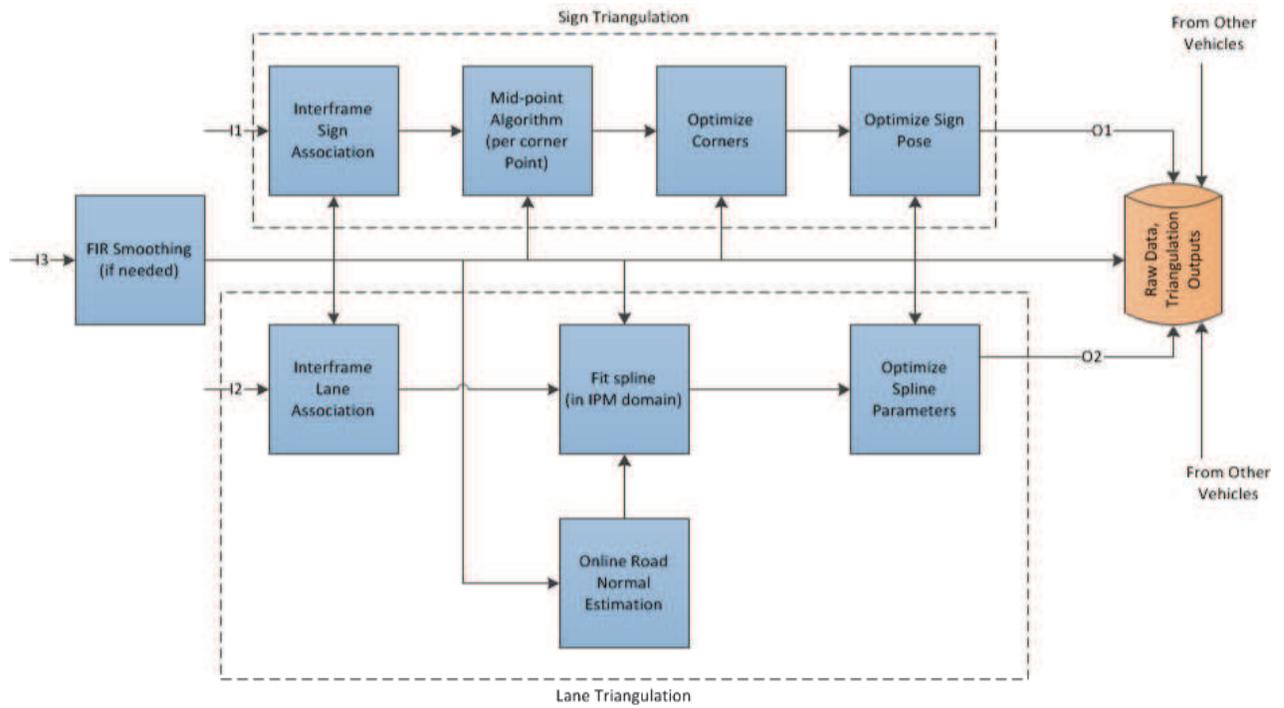}
\caption{Single journey mapping pipeline}
\label{fig:pipeline-sj}
\end{figure*}
\begin{figure*}[thpb]
\centering
\includegraphics[scale=0.6]{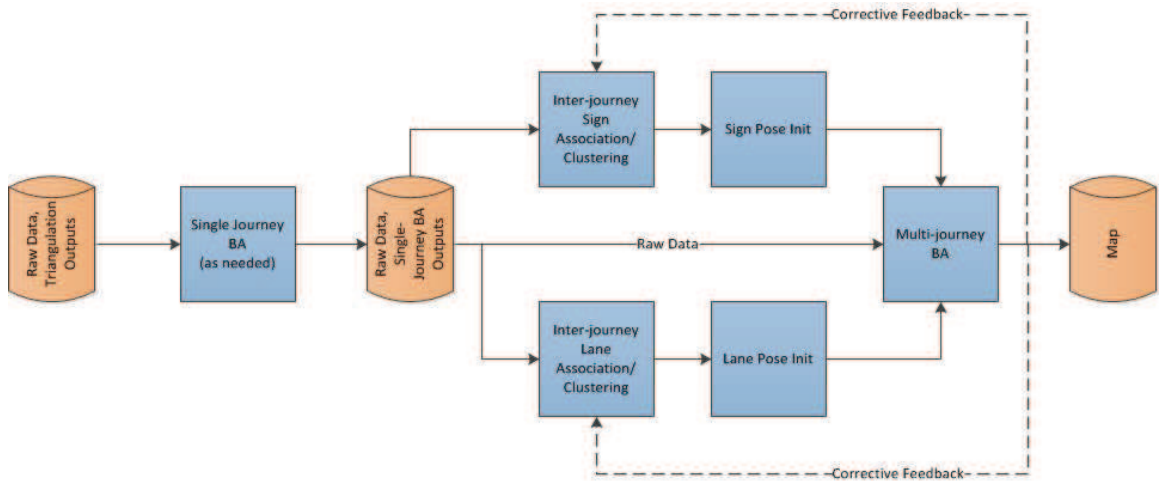}
\caption{Multi-journey mapping pipeline}
\label{fig:pipeline-mj}
\end{figure*}

\subsection{Single-Journey Landmark Triangulation/Reconstruction}
The technical details of this module are given in Section IV. Here we present an overview of some design choices. The single journey processing pipeline receives the following real-time inputs:
\begin{itemize}
\item Traffic sign vertex coordinates in the camera frame and sign shape type for any sign detected or tracked in the frame (denoted by I1 in Fig. \ref{fig:pipeline-sj}).
\item Traffic lane point coordinates in the camera frame for any lane detected in the frame (denoted by I2 in Fig. \ref{fig:pipeline-sj}).
\item Camera 6dof pose for each frame of the camera (denoted by I3 in Fig. \ref{fig:pipeline-sj}).
\end{itemize}
Single-journey triangulation processes these inputs and generates 6dof traffic sign pose in world coordinates (O1 in Fig. \ref{fig:pipeline-sj}) and 3d spline fit for lane segments (O2 in Fig. \ref{fig:pipeline-sj}). In our system, we made the choice of implementing this module in real-time since 3d reconstruction of traffic signs and lanes is also needed by autonomous vehicles for a semantic understanding of their environment. However, for a pure mapping system, an offline implementation is also acceptable.

The sign and lane triangulation pipelines are independent. In both cases we first use the camera pose and the landmark detections for inter-frame associations. In our system, perception module also performs tracking. For modularity we made the choice that perception tracking relies on image features while inter-frame association in the mapping component relies on camera pose information. Section IV.A covers inter-frame association in detail.

Once we have the tracklets from inter-frame association, we run a more or less classical triangulation pipeline for traffic signs. In particular, we exploit the fact that traffic signs are planar. The details of single-journey sign reconstruction are described in Section \ref{sec:sj-signtr}.

For lanes, the situation is more complex. The inverse perspective mapping (IPM) to convert camera lane information onto the road surface in 3d is inherently a sensitive transformation. The fact that lanes are long objects stretching several 10’s of meters ahead of the vehicle further amplifies the problem. One of our key innovations is the estimation of road surface normal near the ego vehicle: we start with a calibrated value but perform periodic online correction of the road normal by ensuring that road points after IPM are planar. The road normal estimation is described in Section \ref{sec:sj-rne} and lane reconstruction is described in Section \ref{sec:sj-lanetr}.

We do not triangulate all the landmarks we detect. To ensure reliability of triangulated landmarks, we apply several pruning rules – for traffic signs often based on the viewing angles and for lanes on the basis of tracklet length.
\subsection{Multi-journey Processing}
To map out a road, we drive on it several times with our vehicles. The real-time triangulation outputs as well as the camera poses and camera frame landmark detections from each of the journeys are stored in the backend mapping server. The aim of the multi-journey processing module is to combine all this information to form the final map. The technical details of this module are described in Section \ref{sec:mj}. Here we give a high level overview as shown in Fig. \ref{fig:pipeline-mj}.

Combining data across multiple journeys can be of two kinds:
\begin{itemize}
\item Jointly process data from multiple journeys so that we are not biased to any particular journey.
\item Incremental update of existing map using new journey data.
\end{itemize}
In practice we need both – the first approach is necessary for cold start (our case) and periodic rationalization of all observed data to ensure an accurate map, while the second approach is necessary for real-time update of the map with new content such as a road construction sign. In this paper we focus on the cold start case.

The first step is to cluster landmarks triangulated from different journeys. For both signs and lanes, we rely on defining suitable distance metrics between the landmarks followed by several rounds of spectral clustering each with different spatial scale parameters. Our current distance metrics are purely based on geometry (including shape of traffic signs), though in future we envision using some underlying image features. Once we have a distance/similarity metric, there is a rich set of clustering algorithm choices. We use spectral clustering since an analysis of similarity matrix eigenvalue spectrum and the spectral gap gives us a good estimate of the number of clusters, that is, the number of true landmarks underlying our data. Once we have estimated the number of landmarks in the map, we either use K-means clustering of the similarity matrix rows or the sign of the row entries for deriving a cluster binary code. The details of multi-journey clustering are given in Section \ref{sec:mj-cluster}.

The final step is to use classical Bundle Adjustment to jointly optimize camera poses, sign poses, and lane spline parameters across multiple journeys to ensure that the landmark observations in the camera images are well approximated and some regularity constraints are also met. The details of this optimization are given in Section \ref{sec:mj-landmark}.

For cold start, an additional final step may be necessary. In multi-journey clustering, we prefer under-clustering and we ignore small clusters. This ensures that we do not put different landmarks in the same cluster, but it also leads to some observations not being used. To harvest these, once a map is generated, we can feed it back to improve clustering and harvest any unused data. We do not describe this aspect in the paper and instead focus in Section \ref{sec:mj} on landmark clustering and Bundle Adjustment.

\section{NOTATION}

\begin{table}[h]
\caption{Notation}
\label{tab:notation}
\begin{center}
\begin{tabular}{|c||p{6cm}|}
\hline
$f_n$  & Camera frame n \\

$t_n$  & timestamp of frame n \\ 

$p_n$ & 6dof camera pose corresponding to frame n  \\

$s_{n,i}$  & i-th sign detected within frame n \\

$s_{n,i,j}$  & j-th corner of i-th sign detected within frame n \\

$l_{n,i}$  & i-th lane marker detected within frame n \\

$l_{n,i,j}$  & j-th point of i-th lane marker detected within frame n \\

$\mathbf{s}_i$  &  i-th sign tracklet’s final state after inter-frame association and $\left\{s_{n,k}\right\}$ are associated observations\\

$\mathbf{l}_i$  & i-th lane marker tracklet’s final state after inter-frame association and $\left\{l_{n,k}\right\}$ are associated observations\\

$S_i$  & i-th sign (after triangulation of $s_i$) \\

$S_{i,j}$  & 3D coordinates of j-th corner of i-th sign\\

$L_i$  & i-th lane marker (after triangulation of $l_i$)\\

$L_{i,j}$  & 3D coordinates of j-th control point of i-th lane marker\\

$\mathbf{S}_i$  & i-th sign after multi-journey association, $\left\{S_i^{(k)}\right\}$ are associated 3D signs\\

$\mathbf{L}_i$  & i-th lane marker after multi-journey association, $\left\{L_i^{(k)}\right\}$ are associated 3D signs\\
\hline
\end{tabular}
\end{center}
\end{table}

The notation for single journey variables is listed in Table \ref{tab:notation}. For the multi-journey case, we use superscript $(k)$ to denote data from $k$-th journey.

\section{SINGLE JOURNEY TRIANGULATION}
\label{sec:sj}
In this section, we detail sign and lane triangulation using as input 6-dof camera poses and sign/lane detections in pixel domain. The overall flow as detailed in Fig. \ref{fig:pipeline-sj} consists of associating detections across frames to obtain tracklets followed by triangulation.
\subsection{Inter-frame Association}
\label{sec:sj-assoc}
We rely entirely on geometry to associate detections across successive frames since low level image features are not available at this stage of the pipeline at present. Furthermore, geometric considerations were adequate since pose is quite accurate and sign/lane features are relatively well separated. Detections are associated with each other across frames to obtain tracklets. A tracklet is effectively a set of detections for a specific physical object. Inter-frame association is performed causally to appropriately associate new detections either with existing tracklets or to spawn new tracklets. We formulate the association of new detections with existing tracklets as a weighted bipartite graph matching problem. Such a formulation naturally relies on an association cost. We spawn new tracklets for detections that have a high cost of associating with all existing tracklets. 
Let $\brac{\lambda_i}_{i\in I}$ denote the tracklets and $\left\{y_{n,j}\right\}_{j\in J}$ denote the pixel measurements for the $|J|$ objects detected on frame $n$. For each tracklet $\lambda_i$, let $d_i$ denote the gap till the previous measurement, i.e., the last measurement for $\lambda_i$ was received at frame $n-d_i$. Let $C_{n,ij}$ denote the cost of associating measurement $y_{n,j}$ with tracklet $\lambda_i$. For a given measurement, say $y_{n,j}$, we spawn a new tracklet if $\min_{i\in I}C_{n,ij}$ is too high. This is to avoid forcing an association when there is none, e.g., when the detections on the current frame are all new and do not correspond to anything seen thus far. Let $J_N\subseteq J$ denote the subset of measurements for which new tracklets will be spawned. For the remaining measurements, we obtain the optimal assignment to tracklets by solving a weighted bipartite matching problem on the matrix $C_n=[C_{n,ij}]_{i\mid d_i<w,j\in J\setminus J_N}$. Note that we ignore tracklets that have not received measurements for a number of successive frames which we call the association window. This is partly due to computational reasons and partly because purely geometry based association cost metrics become increasingly unreliable as distance between the two poses grows. We update $d_i=0$ if a measurement was associated to tracklet $\lambda_i$ in the current frame, else $d_i \rightarrow d_i+1$. For simplicity, we choose the cost of associating a measurement to a tracklet to be the cost of associating with the most recent measurement for that tracklet. We outline the cost of associating two measurements for signs and lanes below.
The basis of the cost is the usual epipolar constraint, i.e., the cost of associating a point p on frame i and point q on frame j is 
$$
\frac{\left|p^T F_{ij} q\right|^2}{\|F_{ij}^T p\|^2+\|F_{ij} q\|^2}
$$
where $F_ij$ is the fundamental matrix corresponding to the two camera poses. 
\subsubsection{Cost for sign association}
The cost of associating two sign detections is the sum of the costs for each pair of vertices. The matching between vertices itself is obtained by solving another weighted bipartite matching problem. 
\subsubsection{Cost for lane association}
Each lane detection is represented by a point cloud in pixel domain. We simplify it by approximating it with a line. The cost of associating two lane detections is the sum of pairwise epipolar costs for points sampled from the lines. In practice, this elementary approach was quite adequate and was in most cases able to associate detections even across tight clover leaves.
\subsection{Sign Reconstruction}
\label{sec:sj-signtr}
Following association, let $\brac{s_{n,i}}$ denote all detections of the $i^{th}$ sign. Recall that $s_{n,i,j}$ denotes the $j^{th}$ corner of the $i^{th}$ sign. Since the camera poses $\brac{p_n}$ are known, we use the mid-point algorithm \cite{Szeliski2010} to estimate the 3D coordinates of the corners $\brac{\hat{S}_{i,j}}$. Note that these corners are not necessarily co-planar while signs are almost coplanar, unless they are badly damaged. For subsequent processing, we reparametrize the sign to be represented by its center, axis and size. Given the output of the mid-point step above, the center is equal to the centroid of the corners, axis is obtained by determining the best plane fit and size estimates follow easily. We use $\bar{S}_i$ to denote this representation. The final triangulation output,  $S_i$,  is obtained by minimizing a re-projection error based optimization metric with $\overline{S}_i$ serving as the initial value.
\subsection{Road Surface Estimation}
\label{sec:sj-rne}
Mapping lane detections onto the 3D world requires knowledge of the road surface. We describe roads locally by their tangent planes which requires (a) a normal which we call the \lq road normal\rq{}, and (b) an offset. In particular, if $\bh\in R^3$  denotes the vector from the camera center to the road plane along the road normal expressed in the camera frame, then $\norm{\bh}$ is the offset and $\bh/\norm{\bh}$ the road normal. This approach results in the road surface being approximated by the convex hull of  a sequence of planes that are tangent to the road surface at the camera location. Recall that the camera is rigidly mounted on the vehicle. Since the vehicle frame is nearly rigidly related to the road surface that it drives on, the road-normal when expressed in the camera frame is nearly constant. This is the premise behind the approach outlined below to estimate the road normal. We measure h during offline calibration, and continuously adjust our estimate of $\bh/\norm{\bh}$ during a drive. While the offset, $\norm{\bh}$, can also change during a drive, we ignore it since inverse perspective projection is far less sensitive to the offset than it is to the road normal. In what follows, road normal always refers to its representation in the camera frame. Road normal estimation is carried out in two steps, (1) offline calibration to get a good initialization, and (2) online adjustment during a drive.
\subsubsection{Offline calibration}
We use an offline calibrated road normal value for initialization of our online adjustment. The setup for this comprises of a wall and floor in our garage with clearly marked points (10 on the wall and 6 on the floor). We use a total station to measure all these points accurately in a local 3d coordinate system. Moreover, using the ground points, we estimate the road normal in local 3d coordinates. Given a camera mounted in a vehicle, we take several pictures of the wall and annotate the wall points in the image after appropriate undistortion. Using the image points and the wall point coordinates in the local frame, we use the ePnP algorithm \cite{Lepetit2008} to estimate the 6dof local frame to camera frame transformation. This transformation is then used to convert the road normal to the camera frame coordinates.
\subsubsection{Online adjustment}
The road normal can change during a drive due to several reasons, e.g., it is sensitive to vehicle suspension, and hence changes by a few degrees with passenger/load configuration, and even a miniscule movement of the camera due to vibrations during a drive can cause an appreciable change in the road normal. Experiments showed that mean absolute lane width estimation errors reduce by about 50\% with online adjustment.  We setup online adjustment as a non-linear least squares problem. We rely on the fact that the horizon does not change much in the image in an automotive use case. Recall that the camera height (measured along the calibrated road normal) from offline calibration is $\norm{\bh}$. We start by choosing a fixed set of points that lie below the horizon in pixel domain. The rays from camera center through these points are expected to intersect the ground. Hence these points only capture the prior knowledge of the approximate horizon and do not require any knowledge of road segmentation. Let their coordinates in ideal camera frame be $U\triangleq \brac{u_i},u_i = [u_{i,0},u_{i,1},1]^T$. Let the current estimate of the road normal be $\hat{h}$ ($\norm{\hat{h}}=1$). From each camera pose, $p_n$, we use $\hat{h}$ to unproject $U$ to get the set of 3D points in spatial frame, say $W_n \triangleq \brac{w_{n,i}},w_{n,i}\in R^3$. Note that
\begin{equation}
w_{n,i}=c_n+R_nu_i \frac{\norm{\bh}}{\hat{h}^T u_i}
\label{eq:ipm}
\end{equation}
Note that all points in $W_n$ are coplanar by construction. In fact, if $\hat{h}$ is equal to the correct road normal, then the points in $W_n$ will be on the local tangent to the road surface for all n. Furthermore, for all camera poses, say $\mN$, on a planar stretch of the road, the points $W_{\mN}\triangleq \brac{w_{n,i}}_{i,n\in \mN}$ will all be coplanar. But if $\hat{h}$ is incorrect, $W_n$ will not coincide with the road surface for any $n$. While the correct road normal guarantees co-planarity, the converse is certainly not true. This is most easily evident in the following scenario: motion is along camera’s z-axis, and the road is flat with its normal along camera’s y-axis. In this case, any $\hat{h}$ in the xy-plane of the camera frame will result in a planar $W_\mN$. We observed in practice that such coincidences are rare and hence we designed the cost function to capture the co-planarity of $W_\mN$. In words, the cost function is the sum of squared distances of points in $W_\mN$ to their best plane fit, and is given by
$$
J_\mN\bra{\hat{h}} = \sigma_{min}\bra{\sum_{w\in W_\mN}\bra{w-\mu_\mN}\bra{w-\mu_\mN}^T}
$$
where $\mu_\mN$ is the mean of the points in $W_\mN$. The final estimate is given by
\begin{equation}
\hat{h}_{opt} = \argmin_{h}\sum_{\mN}J_{\mN}\bra{h}
\label{eq:rne}
\end{equation}
One can use other surrogates for the smallest singular value such as the harmonic mean of singular values. The latter can be expressed explicitly in terms of the matrix coefficients and hence is computationally cheaper than computing the smallest singular value. The summation in \eqref{eq:rne} is over a number of short segments of the journey, each segment being short enough to be planar. In light of the fact that co-planarity is not a sufficient condition for $\hat{h}$ to be accurate, we admit the solution to the optimization problem in \eqref{eq:rne} only if it is sufficiently close in angle to the value provided by offline calibration. The road normal estimate is updated as above every few tens of meters.

\subsection{Lane Reconstruction}
\label{sec:sj-lanetr}
Let $\hat{h}_n$ denote the estimated road normal at pose $p_n$. Using \eqref{eq:ipm}, one can determine the 3D coordinates $L_{n,i,j}\in R^3$ corresponding to each lane detection point $l_{n,i,j}$. Recall that single journey triangulation is performed online. So, we are interested only in local representations for lanes at this stage. A local representation for the $i^{th}$ lane is obtained by fitting a natural cubic spline with a fixed number of control points to a configurable number of the most recent points in $\brac{L_{n,i,j}}$. The control points obtained from this spline fitting constitute a valid representation since a natural cubic spline is uniquely defined by its control points \cite{Holladay1957}. So, each lane is described by multiple overlapping natural cubic splines. Note that this constitutes a purely geometric representation. Topological structure that includes semantic information such as lane mergers, lane splits, lane types, etc., is to be inferred in the backend during multi-journey triangulation. Several other approaches have been discussed in the literature for geometric and/or topological description of lanes, e.g., clothoids in \cite{Dickmanns1992} which are purely geometric and polylines with 0.5m spacing in \cite{Annieway2011} that are both. In \cite{Betaille2010}, authors describe a road network locally and globally both topologically and geometrically, and propose polylines for local geometric description. We used cubic splines instead with the dual motivation of generating accurate geometric representation locally and computational ease of fitting.

\section{MULTI-JOURNEY MAPPING}
\label{sec:mj}
In this section, we detail the 3D reconstruction of sign and lane landmarks using information that is aggregated from multiple journeys. We first explain the inter-journey association schemes for signs and lanes and then the algorithm for 3D reconstruction of these landmarks. 
We represent our landmarks as follows. A sign $S_i$ is represented as an ordered sequence of 3D sign corners, $\brac{S_{ij},1\leq j\leq C}$, where $C$ is the number of sign corners of the sign face. Lanes are represented as cubic splines with several (usually 5) control points. Lane $L_i$ is given a by a sequence of control points $\brac{L_{ij},1\leq j\leq C_L}$, where $C_L$ is the number of controls points. 
We perform association across multiple journeys for signs and lane independently. We describe these algorithms in the following subsections.

\subsection{Multi-journey Landmark Clustering}
\label{sec:mj-cluster}
The triangulated landmarks from individual journeys and their associated positioning and geometric observations are stored in the backend mapping server. The aim of multi-journey sign/lane clustering is to identify the group of triangulation outputs for several individual journeys (from different cars, across different days) that correspond to the same true underlying landmark object. We pose this as a clustering problem, and specifically, we use spectral clustering since it is also able to estimate the number of clusters, that is, the number of true landmarks underlying the triangulation outputs from multiple journeys. Broadly, given any two landmarks, we have a similarity metric defined between them. This metric takes values in $[0,1]$, with a value of 1 indicating perfect match, while 0 implying different objects. Spectral clustering relies on the eigenvalues and eigenvectors of the similarity matrix/associated probability transition matrix or Laplacian to form the clusters \cite{Luxburg2007}. We do not cover its details since it is a well-known technique, however, the intuition is not hard to see. In a perfect noiseless world, if we have exactly $P$ distinct landmarks, then the similarity matrix can be expressed as a block diagonal matrix, where each of the $P$ diagonal blocks is the all ones matrix. Such a matrix has eigenvalue 1 repeated $P$ times and all other eigenvalues are zero. In the real world, we find that in spite of non-idealities, we often do see a clear spectral gap, which allows to estimate the number of clusters well. Once the number of clusters is identified, then a $K$-means clustering of the rows of the Laplacian eigenvector matrix yields the final clusters. In the following sub-sections we describe the specific similarity metrics used for traffic signs and lanes.

\subsubsection{Sign clustering across multiple journeys}
\label{sec:mj-cluster-signs}
At present, visual features such as image patches are not available in the backend to perform data association across the different triangulated signs from multiple journeys, though in future we may add some such features in our system. We only use geometric information about the triangulated signs and find that it is good enough for the highway and suburban roads we have driven on. 

In the multi-journey setting, we introduce superscript $(k)$ notation to denote $k^{th}$ journey associated with a landmark. To perform data association via spectral clustering, we only use the center of the sign face $S_i^{(k)}$, denoted by $\overline{S_i^{(k)}}$. In our experience, we found the sign center is less sensitive to triangulation noise for the clustering process. The distance between two sign triangulations, $S_i^{(k)}$ and $S_{i'}^{(k')}$ is the $L^2$ distance between the sign centers $d\bra{S_i^{(k)},S_{i'}^{(k')}}=\norm{\overline{S_i^{(k)}}-\overline{S_{i'}^{(k')}}}$.
To map the distance metric to similarity value in $[0,1]$, we use a Gaussian kernel to modulate the distance metrics.
\begin{equation*}
Sim\bra{S_i^{(k)},S_{i'}^{(k')}}=exp\bra{-\frac{1}{2}\bra{\frac{d\bra{S_i^{(k)},S_{i'}^{(k')}}}{d_c}}^2}
\label{eq:simsign}
\end{equation*}
where $d_c$ is a tunable critical distance parameter that implicitly controls the clustering radius. Larger $d_c$ will cluster sign triangulations in larger geographical area, and smaller $d_c$ will cluster sign triangulations in a smaller geographical area. We encode the prior that triangulations from a single journey are distinct by initializing $Sim\bra{S_i^{(k)},S_{i'}^{(k')}}=0$. Note that this is a soft constraint and cannot enforce that signs from the same journey aren’t clustered.

Once the similarity matrix is computed, we can perform spectral clustering to cluster the sign triangulations to different clusters. We observed that by setting $d_c = 4m$, we can cluster signs from different sign posts easily, but the clustering algorithm had difficulty in separating signs in a given signpost. Using a smaller $d_c= 1m$, created many clusters yielding several duplicate signs even after clustering. To address this problem, we performed two stage hierarchical clustering in a top-down fashion: the 1st stage clustering is with $d_c = 4m$, and each signs of the clusters from the 1st stage are further cluster using $d_c = 1m$. This yields good clustering and data association. In the multi-journey crowd-sourced framework, we expect a few single journey reconstructed objects $S_i^{(k)}$ to have high errors, which form the outlier points for our clustering approach. To filter such outliers, we discard all cluster that have less than a threshold number of signs. In our experiments, we conservatively only discard sign clusters which have only one sign object, i.e., singleton clusters. We quantify the data loss of this scheme with numerical results in Section \ref{sec:results}.

For each cluster, we defined the cluster-head sign $\bold{S_i}$ obtained by averaging over respective sign-corners of all signs in a given cluster. Thereafter, $\bold{S_i}$ is used as the representative sign to uniquely describe the sign. The sign corners of $\bold{S_i}$ are used in the initialization step of the Bundle Adjustment procedure described in Section \ref{sec:mj-landmark}.

\subsubsection{Lane clustering across multiple journeys}
\label{sec:mj-cluster-lanes}
Recall that lane markers $L_i$, as defined by spline control-points $L_{i,j}$, are obtained from each journey (Section \ref{sec:sj-lanetr}). In this section we describe how lane marker information obtained from different journeys is aggregated. Let  $L_i^{(k)}$  denote the triangulated lane from journey $k$.

Given triangulated lane markers from multiple journeys, parametrized by splines, we want to determine which ones come from the same real-world lane marker. To do this, we define a notion of similarity for a pair of splines and create a similarity matrix by computing it for every pair of splines. We then perform spectral clustering on this matrix using standard techniques. Our main contributions are the definition of the similarity metric and the hierarchical nature of our clustering.

\paragraph*{Similarity metric}
In the triangulation output, our splines are defined by an adaptive number control points depending on the length. To compute the similarity metric, we first sample each spline more finely, having as many sample points as the approximate spline length in meters. Now, for two splines, $L_i$  and $L_j$ with sample points given by the sets $A=\brac{a_i}$ and $B=\brac{b_j}$, we first find the $L^2$ distance between each pair of sample points $a_i$ and $b_j$, $d\bra{a_i,b_j}$. Firstly, the metric $D_{min}\triangleq \min ⁡d\bra{a_i,b_j}$ indicates how close the splines get to each other. Secondly, if we define a threshold $r$, and compute 
$$
N\triangleq \left|\brac{(a_i,b_j ):d(a_i,b_j )<r}\right|/\bra{\left|A\right|\left|B\right|}
$$
This is the (normalized) number of sample point pairs for which the two splines are within a distance $r$ of each other. This indicates the portion for which the splines run alongside each other. Each of these metrics is of interest independently, but our similarity metric combines the two and is computed as $Sim\bra{L_i,L_j }= N/D_{min}$. Additionally, using a Gaussian kernel is also of interest.
\begin{equation}
Sim\bra{L_i,L_j}=exp\bra{-\frac{1}{2}\brac{\frac{D_{min}}{dN}}^2}
\label{eq:simlane}
\end{equation}
where $d$ is an appropriately tuned parameter.

\paragraph*{Hierarchical clustering}
Once the similarity matrix is computed, we compute its eigenvalues, determine the number of clusters, and obtain the clusters by binning the eigenvectors appropriately. Additionally, we have found it useful to cluster hierarchically, that is, instead of creating the desired number of clusters in a single step, we do it in multiple stages, reducing the number of clusters each time, and performing the next round of clustering on the output clusters of the previous round. We typically use two-three stages and find that this gives us better error performance than using only one.
The clustering determines which triangulated lane objects to aggregate together. By creating a point cloud from the control points of these triangulated lanes, and fitting a spline to it, we obtain lane objects $\bold{L_i}$ that are used in landmark reconstruction as described next (Section \ref{sec:mj-landmark}). This serves as the cluster-head lane object of the all lane objects of the given cluster.

\subsection{Landmark Reconstruction}
\label{sec:mj-landmark}
Sign and lane landmarks are reconstructed using a procedure called Bundle-Adjustment (BA) \cite{Agarwal2010} that refines both the 3D location of landmarks and camera poses. The BA primitive is a non-linear optimization that jointly optimizes the camera poses $p_n$ and the landmark locations. It uses camera reprojection error as an error metric for the optimization. Each sign $\bS_i$ has an associated set of image observations. For all frames $f_n$ with poses $p_n$ that have observations of sign $\bS_i$, denoted by $s_{n,i}$, we define the sign reprojection cost to be $\sum_n\norm{\Pi_{p_n}\bra{\bS_i}-s_{n,i}}^2$, where $\Pi_{p_n}$. We use the projection of the sign corners to the image coordinates for pose $p_n$ and the error is computed as the $L^2$ distance between the sign corners in the image coordinates. Each lane $\bL_i$ also has an associated set of image observations. For all frames $f_n$ with poses $p_n$ that contain observations of lane $\bL_i$, we define the lane reprojection cost to be $\sum_n\norm{\Pi_{p_n}\bra{\bL_i'}-l_{n,i}}^2$, where  $\bL_i'$  is the spline point that comes closest to observation $l_{n,i}$ when projected, $\Pi_{p_n}$ is the projection to the image coordinates of that spline point, for pose $p_n$. Then the error is computed as the $L^2$ distance between the sign corners in the image coordinates. The BA optimization problem, then optimizes for \footnote{we also add additional camera pose regularization costs}
\begin{equation}
\argmin_{\brac{p_n,\bS_i,\bL_i}}\bra{\sum_n\norm{\Pi_{p_n}\bra{\bS_i}-s_{n,i}}^2 + \sum_n\norm{\Pi_{p_n}\bra{\bL_i'}-l_{n,i}}^2}
\label{eq:ba}
\end{equation}
Solving \eqref{eq:ba} yields the refined camera poses, sign corners and lane parameters (spline parameters). The problem is a highly non-convex problem and we use iterative Levenberg–Marquardt algorithm \cite{Madsen99} to find a local-minimum. 
The fidelity of the reconstruction is very sensitive to the initialization of the camera pose and landmark parameters. To initialize the landmark parameters $\bS_i$'s and $\bL_i$'s, we use the cluster-heads from the clustering stage. To initialize the camera pose parameters $p_n$’s we aggregated information from multi-journeys to compensate for the observed bias in the single journey camera poses. We inferred the bias by averaging the camera poses and landmark reconstructions across the multiple journeys, and used the bias-compensated camera poses to initialize the $p_n$'s. To make the above optimization problem less sensitive outlier observations, we apply a Tukey weighing function to above cost function to robustify the optimization primitives \cite{Zhang1997}. 

\section{NUMERICAL RESULTS}
\label{sec:results}
\begin{figure}[thpb]
\centering
\includegraphics[scale=1]{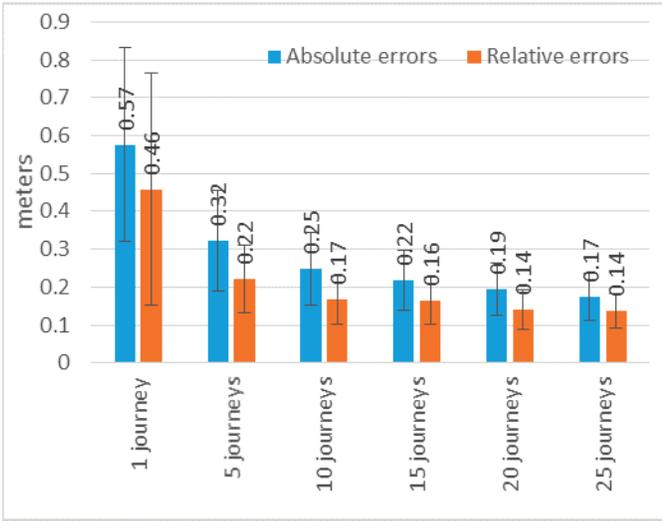}
\caption{Mean sign reconstruction error with varying number of journeys. The error bars correspond to 1-sigma std. dev.}
\label{fig:mj-meansignerror}
\end{figure}
\begin{figure}[thpb]
\centering
\includegraphics[scale=1]{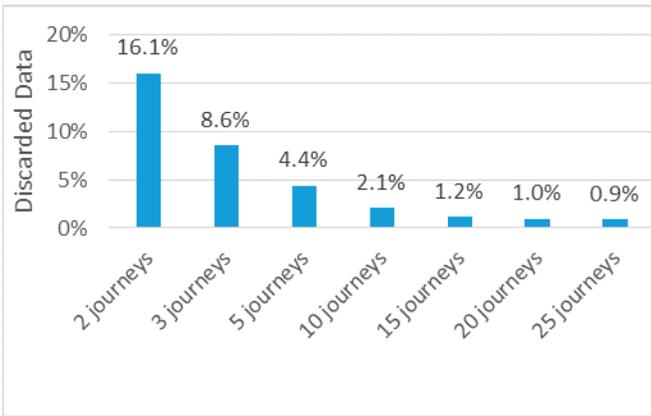}
\caption{Data loss in clustering}
\label{fig:mj-dataloss}
\end{figure}

In this section, we provide some numerical results on the accuracy of sign reconstruction. For evaluation of our mapping solution, we compare our reconstructions with a conventional geological survey using total stations. We have survey results for 31 signs in the surface roads of San Diego, CA, USA. In what follows, we report reconstruction errors relative to these surveyed signs. We analyze reconstruction performance as a function of the number of journeys.

In order to get error metrics, we first associate reconstructed sign corners with surveyed sign corners using weighted bipartite graph matching using appropriate geometric metrics (similar to the high level approach in Section \ref{sec:sj-assoc}). After matching, we compute the absolute error in the sign corners between the reconstructed and surveyed signs. Fig. \ref{fig:mj-meansignerror} illustrates the mean absolute errors across all signs. As expected, the errors reduce with increasing number of journeys. 

Fig. \ref{fig:mj-meansignerror} shows that with 25 journeys, the error floors to 17cm absolute error and 14cm relative error. In our experiments we observe that refinement with BA provides an improvement of only around 5\% to 10\%, which is within the error margins. 

Our algorithm for sign clustering discards outlier signs of single-journey triangulation by filtering out singleton clusters. We expect data-loss from this filtering approach. Fig. \ref{fig:mj-dataloss} quantities this data loss as a function of the number of journeys used for the clustering approach. The figure shows that with more than 15 journeys, the clustering algorithm is able to better fit to the data discarding singleton outliers that constitute around 1\% of all the data. This is a reflection of the fact that clustering performance improves as the number of journeys increases.

\section{CONCLUSION}
The main message of this paper is that using consumer grade equipment and few tens of journeys it is possible to construct precise 3d maps. This is in contrast to current practice of using dedicated fleets with expensive equipment (such as high resolution lidar and high end GPS/IMU), which is not a scalable approach. In our current experiments, we get mean absolute accuracy of less than 20cm at any sign corner and relative error of about 15 cm from 25 journeys. In addition, BA can provide an additional 5-10\% improvement, but also needs lot more backend computation infrastructure. 

The performance of the mapping component critically depends on the positioning and perception inputs it receives. Due to space constraints, we did not cover those details here, but we hope to do so in other publications.

The mapping component design we presented in the paper is our first design. In future we expect to improve the accuracy of our positioning and perception module further and also add more features to the mapping component (such as use of image features in a sign for better triangulation as well as multi-journey association). We expect these additions will improve the accuracy of our solution further.

\bibliography{main}
\bibliographystyle{IEEEtran}

\end{document}